\documentclass[sigconf,letterpaper,9pt]{acmart}
\acmConference[FAT* '19]{FAT* '19: Conference on Fairness, Accountability, and Transparency}{January 29--31, 2019}{Atlanta, GA, USA}
\acmBooktitle{FAT* '19: Conference on Fairness, Accountability, and Transparency, January 29--31, 2019, Atlanta, GA, USA}
\acmYear{2019}
\copyrightyear{2019}



\usepackage{booktabs} 
\usepackage{array}

\setcopyright{acmcopyright}

\acmDOI{10.1145/3287560.3287595}

\acmISBN{978-1-4503-6125-5/19/01}

\acmPrice{15.00}

\begin{document}

\title{Deep Weighted Averaging Classifiers}

\author{Dallas Card$^1$}
\email{dcard@cmu.edu}
\affiliation{
 $^1$\department{Machine Learning Department}
  \institution{Carnegie Mellon University}
  \streetaddress{5000 Forbes Avenue}
  \city{Pittsburgh}
  \state{Pennsylvania}
  \postcode{15213}
}

\author{Michael Zhang$^2$}
\email{mjqzhang@cs.washington.edu}
\affiliation{
  $^2$\department{Paul G.~Allen School of Computer Science and Engineering}
  \institution{University of Washington}
  \streetaddress{185 E.~Stevens Way NE}
  \city{Seattle}
  \state{Washington}
  \postcode{98195-2350}
}

\author{Noah A. Smith$^{2,3}$}
\email{nasmith@cs.washington.edu}
\affiliation{
  $^3$\institution{Allen Institute for Artificial Intelligence}
  \streetaddress{2157 N.~Northlake Way}
  \city{Seattle}
  \state{Washington}
  \postcode{98103}
}

\renewcommand{\shortauthors}{Dallas Card, Michael Zhang, and Noah A. Smith.}

\begin{abstract}

Recent advances in deep learning have achieved impressive gains in classification accuracy on a variety of types of data, including images and text. Despite these gains, however, concerns have been raised about the calibration, robustness, and interpretability of these models. In this paper we propose a simple way to modify any conventional deep architecture to automatically provide more transparent explanations for classification decisions, as well as an intuitive notion of the \textit{credibility} of each prediction. Specifically, we draw on ideas from nonparametric kernel regression, and propose to predict labels based on a weighted sum of training instances, where the weights are determined by distance in a learned instance-embedding space. Working within the framework of \textit{conformal methods}, we propose a new measure of nonconformity suggested by our model, and experimentally validate the accompanying theoretical expectations, demonstrating improved transparency, controlled error rates, and robustness to out-of-domain data, without compromising on accuracy or calibration.
\end{abstract}

\maketitle


\section{Introduction}

For any domain involving complex, structured, or high-dimensional data, deep learning has rapidly become the dominant approach for training classifiers. Although deep learning is somewhat vaguely defined, we will use it here to refer to any architecture which makes a prediction based on the output of a function involving a series of linear and non-linear transformations of the input representation. While the details of these transformations differ by domain (for example, two-dimensional convolutions are often used for images, whereas sequential models with attention are more common for text), most models for binary or multiclass classification include a final softmax layer to produce a properly normalized probability distribution over the label space. In this paper, we explore an alternative to the softmax, yielding what we call a \textit{deep weighted averaging classifier} (DWAC), and evaluate its potential to deliver equally accurate predictions, while offering greater transparency, interpretability, and robustness.\footnote{Our implementation is available at \url{https://github.com/dallascard/DWAC}}

Despite the success of deep learning as a framework, concerns have been raised about the properties of deep models \cite{szegedy.2013}. In addition to typically requiring large amounts of labeled data and computational resources, the parameters tend to be relatively difficult to interpret, compared to more traditional (e.g., linear) methods.
In addition, some deep models tend to be  poorly calibrated relative to simpler models, despite being more accurate \cite{guo.2017.calibration}. Finally, careful evaluation and an explosion of work on adversarial examples have demonstrated that many deep models are more brittle than test-set accuracies would suggest \cite{goodfellow.2015,hendrycks.2017,nguyen.2015.deep,recht.2018}.

Particularly in light of recent controversy and legislation, such as the General Data Protection Regulation (GDPR) in Europe, there has been rapidly growing interest in developing more interpretable models, and in finding ways to provide explanations for predictions made by machine learning systems. Although there is currently an active debate in the field about how best to conceptualize and operationalize these terms \cite{doshi.velez.2017,guidotti.2018}, recent research has broadly fallen into two camps. Some work has focused on models that are inherently interpretable, such that an explanation for a decision can be given in terms that are easily understood by humans. This category includes classic models that can easily be simulated by humans, such as decision lists, as well as sparse linear models, where the prediction is based on a weighted sum of features \cite{breiman.1984,lakkaraju.2016,lou.2012,tibshirani.1996,ustun.2015.slim,wang.2015}. Other work, meanwhile, has focused on developing methods to provide explanations that approximate the true inner workings of more complex models, in a way that provides some utility to the user or developer of a model beyond what is attainable through more direct means \cite{bastani.2017,lei.2016,lundberg.2017,ribeiro.2016,ribeiro.2018,ribeiro.2018.anchors,selvaraju.2016}. 

In this paper, we propose a method which, like those in the former category, offers an explanation that is transparent (in that the complete explanation is in terms of a weighted sum of training instances), but also explore ways to approximate this explanation by using only a subset of the relevant instances. While this approach retains some of the inherent complexity of typical deep models (in that it is still difficult to explain \textit{why} the model has weighted the training instances as it has for a particular test instance), the mechanism behind the prediction is far more transparent than softmax-based models, and the individual instance weights provide a way for a user to examine the basis of the prediction, and evaluate whether or not the model is doing something reasonable. Similarly, while looking at the nearest neighbors of a test point is a commonly-used heuristic to attempt to understand what a model is doing, that approach is only an approximation for models which map each instance directly to a vector of probabilities.

There is, of course, a long history in machine learning of making predictions directly in terms of training instances, including nearest neighbor methods \cite{cover.1967}, kernel methods, including support vector machines \cite{boser.1992,cortes.1995}, and transductive learners more broadly \cite{vapnik.1998}. The main novelty here is to adapt any existing deep model to make predictions explicitly in terms of the training data using only a minor modification to the model architecture, and arguing for and demonstrating the advantages offered by this approach.

As we will describe in more detail below, we propose to learn a function which maps from the input representation to a low-dimensional vector representation of each input. Predictions on new instances are then made in terms of a weighting of the label vectors of the training instances, where the weights are a function of the distance from the instance to be evaluated to all training instances, in the low-dimensional space. This is closely related to a long line of past work on \textit{metric learning} \cite{xing.2002,weinberger.2006,davis.2007,bellet.2013,kulis.2013}, but rather than trying to optimize a particular notion of distance (such as Mahalanobis distance), we make use of a fixed distance function, and allow the architecture of standard deep models to do the equivalent work for us.
This idea is also related to models which use neural networks to learn a similarity function for specific applications, such as face recognition \cite{chopra.2005} or text similarity \cite{mueller.2016}; here we show how this is a more generally applicable way to train models, and we emphasize the connection to interpretability.

Such an approach comes with distinct advantages:
\begin{enumerate}
\item A precise explanation of why the model makes a specific prediction (label or probability) can be given in terms of a subset of the training examples, rank ordered by distance. Moreover, the weight on each training instance implicitly captures the degree to which the model views the two instances as similar. The explanation is thus given in terms of exemplars or \textit{prototypes}, which have been shown to be an effective approach to interpretability \cite{kim.2014}.
\item In \S \ref{sec:approx}, we show that, in many cases, a very small subset of training instances can be used to provide an approximate explanation with  high fidelity to the complete explanation.
\item In addition, it is possible to choose the size of the learned output representation so as to trade off between performance and interpretability. For example, we can use a lower dimensional output representation if we wish to make it easy to directly visualize the embedded training data.
\item Even in cases where revealing the training data is not feasible, it is possible to provide an explanation purely in terms of weights and labels. Although this does not reveal the \textit{way} in which a new instance is viewed (by the model) as similar to past examples, it sill provides a quantifiable notion of how unusual the new example is. The form of our model suggests a natural metric of nonconformity, and in \S \ref{sec:methods_conformal}, we formalize this using the notion of \textit{conformal methods}, describing how the relevant distances can be used to either provide bounds on the error rate (for data drawn from the same distribution), or robustness against outliers.
\item Finally, although this model does entail a slight cost in terms of increased computational complexity, the difference in terms of speed and memory requirements at test time can be minimized by pre-computing and storing only the low-dimensional representation of the training data (from the final layer of the model). The cost during training will in most cases be dominated by the other parts of the network, and it is still possible to train such models on large datasets without difficulty. Moreover, in our experiments, this choice seemingly involves no loss in accuracy or calibration.
\end{enumerate}

Our \textbf{deep weighted averaging classifiers} (DWACs) are ideally suited to domains where it is possible to directly inspect the training data, such as controlled settings like social science research, medical image analysis, or managing customer feedback.  In these domains, DWACs can create a more transparent and interpretable version of any successfully developed deep learning architecture. Although the advantages are diminished in domains where privacy is a concern, presenting information solely in terms of weights and labels still provides a useful way to quantify the credibility of a prediction, even without allowing direct inspection of the original training data.


\section{Background}

\subsection{Scope and Notation} \label{sec:notation}

In this paper we will be concerned with the problem of classification. In general, we will assume a set of $m$ training instances, $\mathbf{x}_i$ for $i = 1,\ldots,m$, with corresponding labels in some categorical label space, $y_i \in \mathcal{Y}$, for $i = 1, \ldots, m$, where $c = |\mathcal{Y}|$ is the number of classes. We also assume we will eventually be given a set of $n$ test instances, $\textbf{x}_i, y_i$, for $i = 1+m, \ldots, n+m$. We will use $\mathbf{h}_i$ to refer to the output representation of our model for instance $i$. Square brackets $[\cdot]_k$ will denote the $k$th element of a vector.

\subsection{Nonparametric Kernel Regression} \label{sec:nw}

We propose to build on a classic method from nonparametric regression, known as Nadaraya-Watson (NW). The original use case of NW was in regression, with $y_i \in \mathbb{R}$, where it is assumed that $y_i = m(\textbf{x}_i) + \varepsilon$, where $\mathbb{E}[\varepsilon] = 0$ and $\mathbb{V}[\varepsilon] = \sigma^2$. The goal is to estimate the mean function, $m(\textbf{x})$, which can be expressed in terms of the joint density as
\begin{align} 
m(\textbf{x}) = \mathbb{E}[Y \mid X = \textbf{x}] = \int y~P(y \mid \textbf{x})~dy = \frac{\int y~P(\textbf{x}, y)~dy}{\int P(\textbf{x}, y)~dy}.
\end{align}

By approximating the joint density using kernel density estimation, it is possible to re-express this as a weighted sum of training instances, i.e., 
\begin{align}
\hat m(\textbf{x}) = \frac{\sum_{i=1}^m y_i~K(\textbf{x}, \textbf{x}_i)}{ \sum_{j=1}^m K(\textbf{x}, \textbf{x}_j)},
\end{align}
where $K(\textbf{x}, \textbf{x}_i)$ is a kernel, such as a Gaussian \cite{nadaraya.1964,watson.1964}.  It is easy to see that this corresponds to a \textit{linear smoother}, in that it will predict outputs as a weighted sum of training instances, i.e.,
\begin{align}
\hat m(\textbf{x}) = \sum_{i=1}^m y_i~\alpha_i(\textbf{x}),
\end{align}
where $\alpha_i(\textbf{x}) = \frac{K(\textbf{x}, \textbf{x}_i)}{\sum_{j=1}^m K(\textbf{x}, \textbf{x}_j)}$.

This method can easily be adapted to classification by predicting the \textit{probability} of an output label as a weighted sum of training labels, i.e., 
\begin{align}
P_\textrm{NW}(y =k \mid \textbf{x}) = \sum_{i=1}^m \mathbb{I}(y_i = k)~\alpha_i(\textbf{x}),
\end{align}
where $\mathbb{I}(\cdot)$ equals 1 if the condition holds or 0 if not.

The primary limitation on NW is due to the curse of dimensionality: as the dimensionality of $\textbf{x}$ grows, sparse data becomes a problem, and the notion of proximity becomes problematic \cite{aggarwal.2001}.
In this work, we show how we can avoid this problem by using neural networks to embed inputs into a space with reasonable dimensionality, and proceed to compute weights over training instances in that space.

\subsection{Conformal Methods} \label{sec:conformal}

\textit{Conformal methods} refer to a broad set of ideas which aim to provide theoretical guarantees on error rates in classification or regression \cite{saunders.1999,vovk.2005,shafer.2008,lei.2011}. Conformal methods can be used with any base classifier or predictor, and work by introducing a generic notion of \textit{nonconformity}. As will be explained in detail below, each possible prediction (i.e., label or value) that can be made for a given test instance can be evaluated in terms of its nonconformity. By comparing these with the nonconformity scores of either all data in a leave-one-out manner, or with a held-out \textit{calibration} set, the equivalent of a $p$-value can be associated with each possible prediction, allowing for thresholding in a way that provides a guarantee on the error rate (for i.i.d. or exchangeable data).

Because of the high computational cost of the leave-one-out approach to conformal predictors, we will focus here on the approach based on a held-out calibration set.\footnote{The leave-one-out approach may be superior in terms of statistical efficiency, but it is computationally infeasible for all but very small datasets.} In particular, we will begin by shuffling our training data, and partitioning it into a proper training set ($i=1,\ldots,t$), and a calibration set ($i=t+1,\ldots,m$).\footnote{Our calibration set will also serve as our validation set for early stopping during training.}

The fundamental choices in conformal methods are a base classifier and a measure of nonconformity, $A$. The latter concept, which intuitively corresponds to how \textit{atypical} an instance is, maps a bag of examples (i.e., the proper training set, with labels), and one additional instance ($\mathbf{x}$, the observed features) with one possible label, $k \in \mathcal{Y}$, to a scalar $\eta \in \mathbb{R}$, i.e.,
\begin{align}
\eta(\mathbf{x}, k) = A \left(  \Lbag (\mathbf{x}_i, y_i) \Rbag_{i=1}^t, (\mathbf{x}, k) \right),
\end{align}
where $\Lbag \cdot \Rbag$ denotes a \textit{bag}, i.e., a multi-set (potentially containing duplicate instances).

The idea of a measure of nonconformity is quite general, but in practice, the most common approach is to convert a bag of examples into a \textit{model}, and then compare the \textit{prediction} of that model on the training instance ($\mathbf{x}$) with the hypothesized label, $k$. In particular, for any probabilistic model, the simplest measure of model-based nonconformity is any inverse monotonic transform of the predicted probability of the hypothesized label, such as the inverse or negation. For example, the following is a valid measure of nonconformity:
\begin{align}
\eta(\mathbf{x}, k) = - P_{\mathcal{M} \left( \Lbag (\mathbf{x}_i, y_i) \Rbag_{i=1}^t \right)}\left( y=k \mid \mathbf{x} \right),
\end{align}
where $\mathcal{M} \left( \Lbag (\mathbf{x}_i, y_i) \Rbag_{i=1}^t \right)$ represents a model trained on the proper training set, $i=1,\ldots,t$.

Conformal methods work by comparing the nonconformity score of each possible prediction for each test instance to the nonconformity scores of all instances in the calibration set. Specifically, we will compute a $p$-value for each hypothetical test instance label equal to the proportion of the calibration instances that have a higher nonconformity score (i.e., a value between 0 and 1 indicating how \textit{conforming} a possible label for a new instance is, relative to the calibration data). More precisely, for a test instance $\mathbf{x}$ and hypothesized label, $k$, 
\begin{align}
p(\mathbf{x}, k) = \frac{\sum_{j=t+1}^m \mathbb{I} \left( \eta(\mathbf{x}_j, y_j) \geq \eta(\mathbf{x}, k) \right)}{m - t},
\end{align}
where $\eta$ is computed relative to the proper training set for all instances (both calibration and test).

Unlike traditional classifiers, conformal methods may predict anywhere from zero to $c$ labels for a given instance.  We will revisit this choice below, but for the moment, the decision rule will be that for each test instance, we make a positive prediction for all labels $k$ for which $p(\mathbf{x}, k) > \varepsilon$, where $\varepsilon \in [0, 1]$ is chosen by the user. By the properties of conformal predictors, these predictions will be asymptotically valid; that is, both theoretically and empirically, for i.i.d. data, this will produce a set of predicted labels for each test instance such that at least $1-\varepsilon$ of the predicted label sets include the true label, with high probability. Moreover, this property holds for any measure of nonconformity.\footnote{More precisely, for any measure of nonconformity, and i.i.d. data drawn from any distribution, the unconditional probability that the true label of a random test instance is not in the predicted label set does not exceed $\varepsilon$ for any $\varepsilon$. The broader statement follows from the law of large numbers. It is also possible to bound the conditional probability of error (conditional on the training data) with slightly weaker guarantees. For more details, please refer to \cite{vovk.2005,vovk.2012}.}  Of course, this property is trivially easy to satisfy by  predicting all labels for all test instances. In practice, however, better choices of measures of nonconformity will be more or less \textit{efficient}, in that they will tend to produce smaller  predicted label sets without compromising on error rate.

In addition, for any given test instance, we can compute two properties of the label that is predicted with the highest probability, which Saunders et al. called \textit{confidence} and \textit{credibility} \cite{saunders.1999,shafer.2008}. Confidence is equal to the largest $1-\varepsilon$ such that the predicted label set includes only a single label (i.e., one minus the second-largest $p$-value among the possible labels). This corresponds to the probability, according to the model, that the label which is predicted to be most likely is correct. For example, for i.i.d.~data, we would expect 92\% of predictions with 92\% confidence to be correct, and 8\% to be incorrect.

Credibility, by contrast, is equal to the largest $\varepsilon$ for which the predicted label set is empty (i.e., the largest $p$-value among the possible labels). This correspond to one minus the model's confidence that \emph{none} of the possible labels are correct. In other words, predictions with low credibility indicate that even the most likely prediction is relatively nonconforming compared to the calibration instances (with their true labels), and that we should therefore be skeptical of this prediction.

The idea of not predicting any label is somewhat foreign to conventional approaches, as we can only obtain better accuracy by hazarding at least some guess, and in some applications it would be reasonable to still predict the most probable label according to the model, with an associated confidence and credibility. In other circumstances, however, there may be valuable information in an empty label set. For example, if we set $\varepsilon$ to 0.2, and find that 20\% of the test instances end up with empty predicted label sets, this is a strong indication that the predictions for most of the remaining instances are correct (because the overall error rate is expected to be 20\%). In practice, this would be an unusual outcome, but we will explore the usefulness of this sort of signal in the context of outlier detection.


\section{Deep weighted averaging classifiers}

We now turn to our new method, deep weighted averaging classifiers.
\subsection{Model Details}

Most neural network models for classification take the form
\begin{align}
P(y = k \mid \mathbf{x}) = \frac{\exp([\mathbf{h}]_k)}{\sum_{j=1}^c \exp([\mathbf{h}]_j)}, \label{softmax}
\end{align}
where $\mathbf{h} = \mathbf{W} \cdot f(\mathbf{x}) + \mathbf{b}$. In this architecture, $f(\mathbf{x})$ embeds the input in a model-specific way (e.g., a convolutional or recurrent network) into a lower-dimensional vector representation.
Although $\mathbf{W}$ and $\mathbf{b}$ could be folded into $f(\mathbf{x})$, we make them explicit to emphasize that $\mathbf{W}$ is required to project the output of $f(\mathbf{x})$ (which is a vector of arbitrary length) down to a vector of length $c$ (the number of classes).
The softmax (Eq.~\ref{softmax}) then projects this vector onto the simplex.

Our proposed alternative is to leave $f(\mathbf{x})$ unchanged, eliminate the softmax, and to redefine the predicted probability as 
\begin{align}
P(y = k \mid \mathbf{x}) = \frac{\sum_{i=1}^t \mathbb{I}(y_i = k)~w(\mathbf{h}, \mathbf{h}_i)}{ \sum_{j=1}^t w(\mathbf{h}, \mathbf{h}_j) },
\end{align}
i.e., a weighted sum over the labels of the instances in the proper training set, where $\mathbf{h}$ is defined as above and $w(\mathbf{h}, \mathbf{h}_i)$ is a function of the similarity between the embeddings of $\mathbf{x}$ and $\mathbf{x}_i$ in the low-dimensional space, according to some metric. In this architecture, the dimensionality of $\mathbf{h}$ is arbitrary, and the size of $\mathbf{W}$ and $\mathbf{b}$ can be modified as necessary.\footnote{We could of course dispense with $\mathbf{W}$ and $\mathbf{b}$ here, and compute $w(\cdot, \cdot)$ directly on the output of $f(\mathbf{x})$, but we wish to remain as close as possible to the softmax model for the purpose of comparison, while allowing for the possibility of varying the size of $\mathbf{h}$.}

An obvious choice of weight function is a Gaussian kernel operating on Euclidean distance, i.e.,
\begin{equation} \label{eq:gauss}
w(\mathbf{h}, \mathbf{h}_i) = \exp \left( \frac{- \| \mathbf{h} - \mathbf{h}_i\|_2^2}{2 \sigma} \right).
\end{equation}

Typically, in using NW, or other kernel smoothers, one needs to choose the bandwidth, equivalent to $\sigma$ in equation (\ref{eq:gauss}). However, because we will assume that we will be building this classifier on top of a high-capactity embedding network, $f(\mathbf{x})$, we will simply fix $\sigma = \frac{1}{2}$, and force $f(\mathbf{x})$ to adapt to this distance function.

\subsection{Training}

In order to learn $\mathbf{W}$, $\mathbf{b}$, and all parameters of $f(\mathbf{x})$, we use stochastic gradient descent to optimize the log loss on the training data. Because it is impractical to compute the exact probabilities according to the model during training (because they depend on all training instances), we instead rely on an approximation based on the other instances within each minibatch. Specifically, on each epoch of training, we shuffle all instances in the proper training set into minibatches of size $B$. For each minibatch, $\mathcal{B}$, we then minimize 
\begin{align} \label{eq:loss}
\mathcal{L}(\mathcal{B}) = \frac{1}{B} \sum_{j \in \mathcal{B}} \sum_{k=1}^c - \mathbb{I}(y_j=k) \log \hat P(y_j = k \mid \mathbf{x}_j),
\end{align}
where  
\begin{align}
\hat P(y_j =  k \mid \mathbf{x}_j) = \frac{\sum_{i \in \mathcal{B} \backslash \{j\}} \mathbb{I}(y_i = k)~w(\mathbf{h}_j, \mathbf{h}_i)}{ \sum_{l \in \mathcal{B} \backslash \{j\}} w(\mathbf{h}_j, \mathbf{h}_l) }.
\end{align}
As usual, the loss function in equation (\ref{eq:loss}) can be augmented with a regularization term if desired.

Although computing all pairwise distances between many points is relatively expensive, this can be done efficiently for minibatches using standard matrix operations on a GPU.
Specifically, a forward pass through the last layer of a DWAC model (i.e., computing probabilities for one minibatch) requires $\mathcal{O}(B^2h)$, where $B$ is the size of the minibatch and $h=\textrm{len}(\mathbf{h})$. This will typically be larger than the last layer of the softmax classifier, which is $\mathcal{O}(Bhc)$, where $c$ is the number of classes, but in most cases will be dominated by the cost of computing $f(\mathbf{x})$. In practice, the only significant increase in training time is due to the need to embed all training instances in order to estimate performance on a validation set after each epoch. As such, the training runtime will tend to be no worse than twice that of training a softmax model. Similarly, at test time, the computational cost of making a prediction on one test instance is dominated by the cost of embedding the training data. However, this can be pre-computed after training, and only the low-dimensional $\mathbf{h}$ vectors need be stored.\footnote{Note that we only need to compute distances in the low-dimensional space, for which we can choose an appropriately small dimensionality.}

\subsection{Prediction and Explanations} \label{sec:prediction}

Once the model has been trained, predictions can easily be made using the entire training dataset, rather than using a subset, as when computing the loss during training. The complete and explicit explanation for why the model predicts a particular label or probability can then be given explicitly in terms of the training instances, along with the weight on each instance. Moreover, if we consider the closest points (which will be most heavily weighted) as being the most similar, relevant, or important, it is reasonable to provide a sorted list of examples as the explanation. Because the later examples will carry less weight, in many cases only a subset of instances needs to be provided (because for many instances, the lower-weighted training instances will be unable to affect the prediction, no matter what their labels may be).

If we wish to provide an even simpler but approximate explanation, we can also choose to provide only the closest $k$ examples as the explanation, which is a commonly used heuristic for trying to understand model behavior. Although we would not expect that using only a small set of examples would provide a well-calibrated \textit{probability}, it could still provide a reasonable approximate explanation for why the model predicted a particular \textit{label}, assuming that there is strong agreement between predictions made using such a subset and the full model (which we will empirically evaluate in the experiments below). 

\subsection{Confidence and Credibility} \label{sec:methods_conformal}

As discussed in section \S \ref{sec:conformal} above, for any probabilistic classifier, we can use any monotonic transformation of the predicted probabilities which reverses their order (such as $-P(\mathbf{x})$ or $1/P(\mathbf{x})$) as a valid measure of nonconformity. However, our proposed architecture suggests another measure, namely the negated unnormalized weighted sum of training labels of the hypothesized class, i.e.,
\begin{align} \label{eq:measure}
\eta(\mathbf{x}, k) = A\left( \Lbag (\mathbf{x}_i, y_i) \Rbag_{i=1}^t , (\mathbf{x}, k ) \right) = -\sum_{i=1}^t \mathbb{I}(y_i = k)~w(\mathbf{h}, \mathbf{h}_i).
\end{align}

Because of the properties of conformal methods, this measure of nonconformity is automatically \textit{valid}. It may, however, be more or less \textit{efficient} than other measures, such as ones based on probabilities. We note, however, that this proposed measure has an intuitive explanation in terms of how close the training points of the predicted class are (in the embedded space) to the instance for which we wish to make a prediction. Naturally the absolute distance has no meaning in the embedded space, but we avoid this problem by scaling the measure of nonconformity relative to calibration data in order to obtain a $p$-value, as is always the case in conformal methods.

When using probability as the basis of nonconformity, the farther a point is from the decision boundary, the higher will be its predicted probability, and therefore its credibility. Under the measure we propose in equation (\ref{eq:measure}), by contrast, predictions will only be associated with high credibility when the embedded representation of that instance is relatively close to the embedded training instances. That is, if we encounter an instance that is unlike anything seen in the training data, and if the model embeds that instance such that it is far away from all embedded training instances, then this measure will tell us that it is highly nonconforming for all classes, which will result in the model's prediction having very low credibility. As we show below, this is a useful way to quantify the degree to which we should be skeptical of the model's prediction.


\section{Experiments}

To demonstrate the potential of DWAC models, we provide a range of empirical evaluations on a variety of datasets. In addition to showing that our proposed approach is capable of obtaining equivalently accurate predictions, we also compare to a baseline in terms of calibration and robustness to outliers, and illustrate the sorts of explanations offered by a DWAC model. We also empirically validate that the theoretical guarantees claimed by conformal methods hold for both the softmax and the DWAC model, both with and without our proposed measure of nonconformity.

In all cases we report accuracy and calibration (i.e., the accuracy of the predicted probabilities), measuring the latter in terms of mean absolute error (MAE), using the adaptive binning approach of Nguyen and O'Connor \cite{nguyen.2015}. For the initial comparison between models, we only consider conventional prediction (i.e., only using the top-scoring label predicted by each model), and separately evaluate conformal prediction in \S\ref{sec:conformal_results}. Note that our purpose here is not to demonstrate state-of-the-art performance on any particular task, but rather to show that our proposed modification works for a wide variety of architectures.
 
In order to evaluate robustness to outliers, we consider two approaches. The first is to drop one class from a multiclass dataset, and treat the held-out class as out-of-domain data. The other approach is to find a dataset that has a similar input representation, but is fundamentally different in terms of content, and again, treat these instances as out-of-domain.

\subsection{Datasets}

For our experiments we make use of datasets of three different types (tabular, image, and text), including both binary and multiclass problems. For tabular data, we use the familiar Adult Income \cite{kohavi.1996} and Covertype \cite{blackard.1999} datasets available from the UCI machine learning repository, as well as the Lending Club loan dataset available through Kaggle. For images we use CIFAR-10 \cite{cifar.2009} and Fashion MNIST \cite{xiao.2017}. For text we use paragraph-length product reviews (Amazon; 1--5 stars) \cite{mcauley.2015} and movie reviews (IMDB; positive or negative) \cite{maas.2011}, sentences extracted from movie reviews and labeled in terms of \textit{subjectivity} \cite{pang.2004}, and a dataset of Stack Overflow question titles sampled from 20 different categories \cite{xu.2015}. 
Table \ref{tab:datasets} summarizes the most important properties of these datasets; more details and URLs can be found in supplementary material.

\begin{table}
\begin{tabular}{l c c c c}
Dataset & Type & \# classes & \# instances & \# features \\
\hline
Adult Income & Tabular & 2 & 45000 & 103 \\
Covertype & Tabular & 7 & 581000 & 54 \\
Lending Club & Tabular & 2 & 626000 & 157 \\ 
Fashion MNIST & Image & 10 & 60000 & 784  \\
CIFAR-10 & Image & 10 & 60000 & 3072 \\
Subjectivity & Text & 2 & 10000 & 20000 \\
Stack Overflow & Text  & 20 & 20000 & 15000 \\
IMDB & Text & 2 & 50000 & 112000 \\
Amazon & Text & 5 & 190000 & 144000 \\
\end{tabular}
\caption{Properties of datasets used in this paper. For text data, we report vocabulary size as the number of features.}
\label{tab:datasets}
\end{table}

\subsection{Models and Training}

In all cases we choose a base model appropriate to the data. For the tabular data, we use a simple three-layer multi-layer perceptron. For images, we use multi-layer convolutional neural networks. For text datasets, we use a shallow convolutional model with attention \cite{mullenbach.2018}.
In all cases we compare DWAC and softmax models of equivalent size, but also explore varying the dimensionality of $\mathbf{h}$ in the DWAC model. We use the standard train/test split where available, and otherwise sample a random 10\% of the data for a test set, and always use a random 10\% of the training data as a validation/calibration set. For measuring accuracy and calibration on test data, we average over 5 trials with different splits of the training data into a proper training set and a validation/calibration set, with the same split being given to both models. For both models we use Adam \cite{kingma.2014} with an initial learning rate of 0.001 and early stopping. For more details, please refer to supplementary material.


\section{Results}

\subsection{Classification Performance}

As shown in Table \ref{tab:results}, except for one dataset (Covertype) the performance of our model is indistinguishable from a softmax model of the same size in terms of accuracy. As noted above, these results are based on the single best-scoring label from each model, without yet incorporating the idea of conformal prediction. For calibration, the DWAC model is sometimes slightly better and sometimes slightly worse, although we note in passing that, at least for these models and datasets, the predictions from both models are quite well calibrated, such that the predicted probabilities are relatively reliable, at least for in-domain test data. As expected, runtime during training was approximately 50\% longer per epoch than for the equivalent softmax model, with a similar number of epochs required.

\begin{table}
\centering
\begin{tabular}{l c c}
 & \multicolumn{2}{c}{Accuracy $\uparrow$}\\
Dataset & Softmax & DWAC \\
\hline
Adult Income & 0.851 ($\pm$0.002) &  0.850 ($\pm$0.002) \\
Covertype & 0.774 ($\pm$0.003) & 0.760 ($\pm$0.001) \\
Lending Club & 0.956 ($\pm$0.001) &  0.955 ($\pm$0.001) \\
Fashion MNIST  & 0.928 ($\pm$0.002) & 0.927 ($\pm$0.002) \\
CIFAR-10 & 0.898 ($\pm$0.004) & 0.897 ($\pm$0.009) \\
Subjectivity & 0.948 ($\pm$0.002)  & 0.946 ($\pm$0.004) \\
Stack Overflow & 0.869 ($\pm$0.005) & 0.866 ($\pm$0.008) \\
IMDB & 0.905 ($\pm$0.002) &  0.904 ($\pm$0.001) \\
Amazon & 0.740 ($\pm$0.002) & 0.738 ($\pm$0.002) \\
\hline\hline
 & \multicolumn{2}{c}{Calibration (MAE) $\downarrow$} \\
Dataset & Softmax & DWAC \\
\hline
Adult Income & 0.012 ($\pm$0.006) & 0.018 ($\pm$0.002)  \\
Covertype & 0.005 ($\pm$0.001) & 0.010 ($\pm$0.001) \\
Lending Club & 0.007 ($\pm$0.001) & 0.014 ($\pm$0.001) \\
FashionMNIST & 0.006 ($\pm$0.001) & 0.003 ($\pm$0.001) \\
CIFAR-10 & 0.011 ($\pm$0.001) &  0.009 ($\pm$0.002) \\
Subjectivity & 0.020 ($\pm$0.006) & 0.023 ($\pm$0.006) \\
Stack Overflow & 0.009 ($\pm$0.001) & 0.010 ($\pm$0.001) \\
IMDB & 0.029 ($\pm$0.006) & 0.024 ($\pm$0.010)	 \\
Amazon & 0.008 ($\pm$0.002) & 0.004 ($\pm$0.001) \\
\end{tabular}
\caption{Accuracy (higher is better) and calibration (lower is better) on various datasets using the single best-scoring predicted label from softmax and DWAC models of equivalent size, with standard deviations in parentheses.
}
\label{tab:results}
\end{table}

\begin{figure}
\includegraphics[scale=0.55]{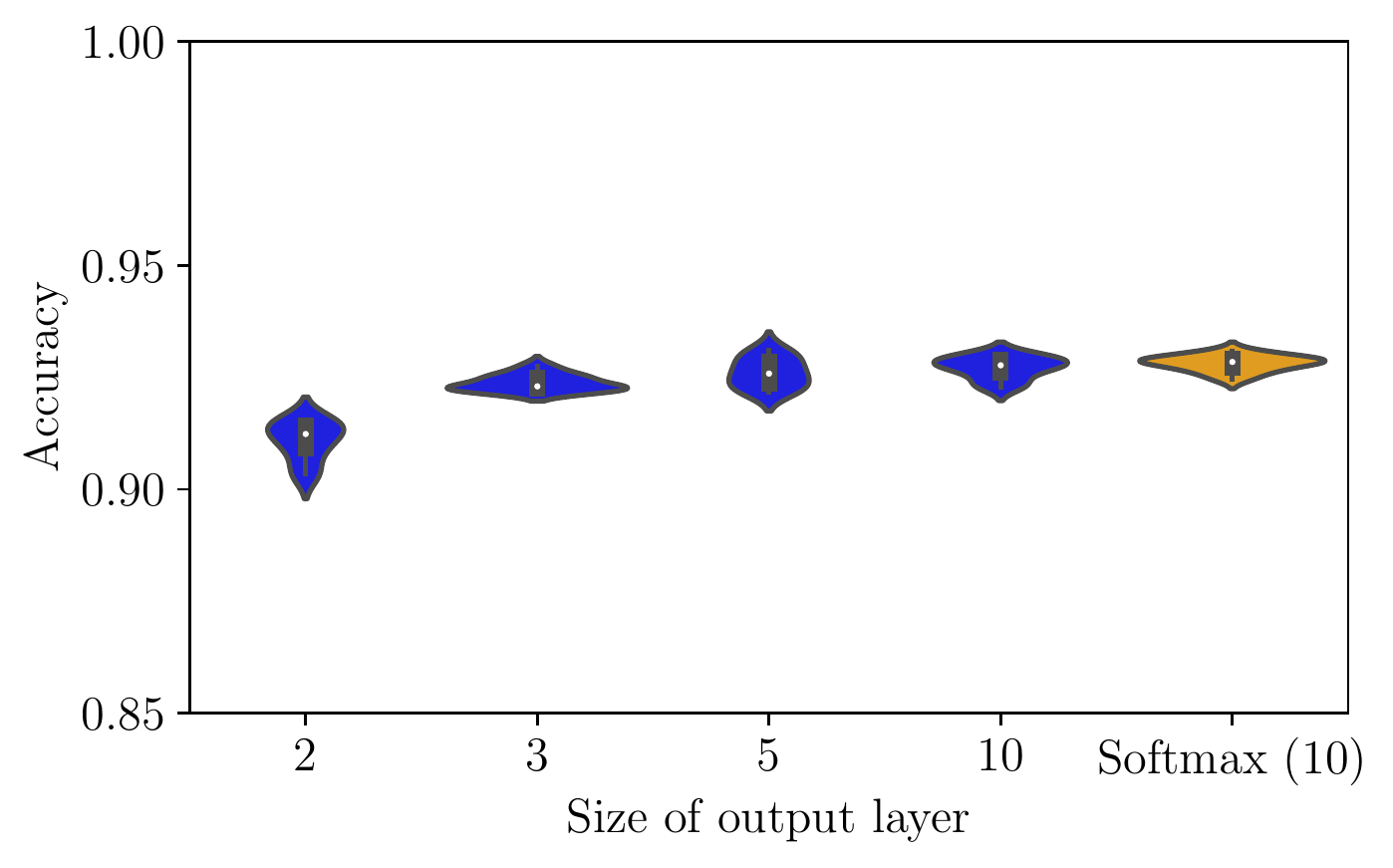}
\caption{Accuracy of DWAC in comparison to a softmax model on the 10-class Fashion MNIST dataset for varying dimensionality of the output layer. Performance is indistinguishable for a DWAC model of the same size, but accuracy drops if we decrease the size of the output layer too much.}
\label{fig:dh}
\end{figure}

One advantage of DWAC is the freedom to choose the dimensionality of the final output layer, and Figure \ref{fig:dh} illustrates the impact of this choice on the performance of our model. While using the same dimensionality as the softmax model gives equivalent performance, the same accuracy can often be obtained using a lower-dimensional representation (i.e., few total parameters). In some cases, however, reducing the dimensionality too much (e.g., two-dimensional output for Fashion MNIST) results in a slight degradation of performance.

On the other hand, using a two-dimensional output layer means that we are able to more easily visualize the learned embeddings, without requiring an additional dimensionality reduction step, e.g., using PCA or t-SNE \cite{vandermaaten.2008}. In this way, we could look directly at where a test instance is being embedded, relative to training instances, with no loss of fidelity. 
\begin{figure}
\includegraphics[scale=0.5]{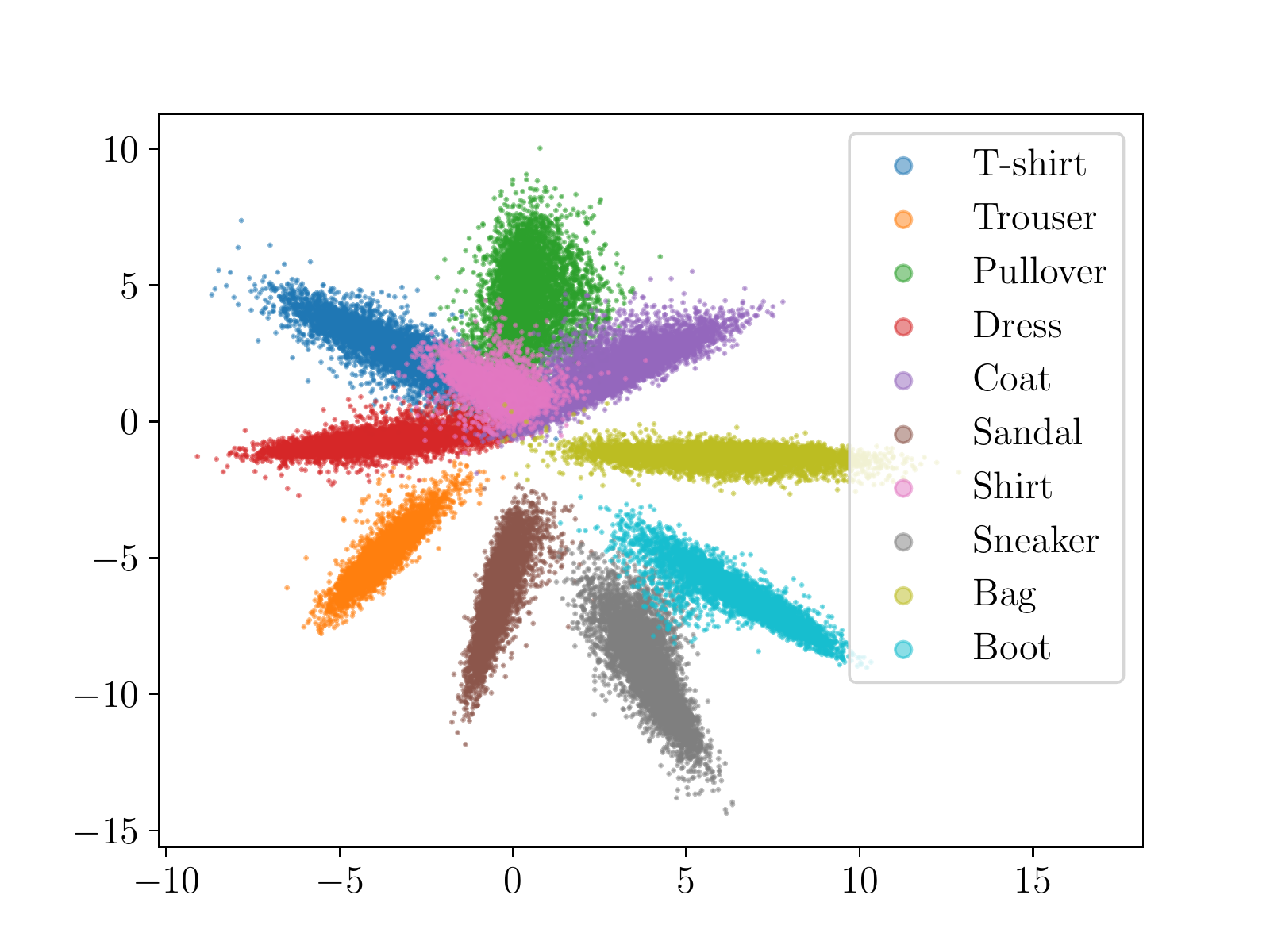}
\caption{Learned embeddings of the Fashion MNIST training data when using a DWAC model with a two-dimensional output layer.}
\label{fig:fashion_scatter}
\end{figure}
Figure \ref{fig:fashion_scatter} shows the embeddings learned by our model for the Fashion MNIST training data using a two-dimensional output layer. Pleasingly, there is a natural semantics to this space, with all of the footwear occurring close together, and the ``shirt'' class being centrally located relative to related classes (t-shirt, pullover, etc.). 

\subsection{Interpretability and Explanations}

\begin{figure}
\includegraphics[scale=0.6]{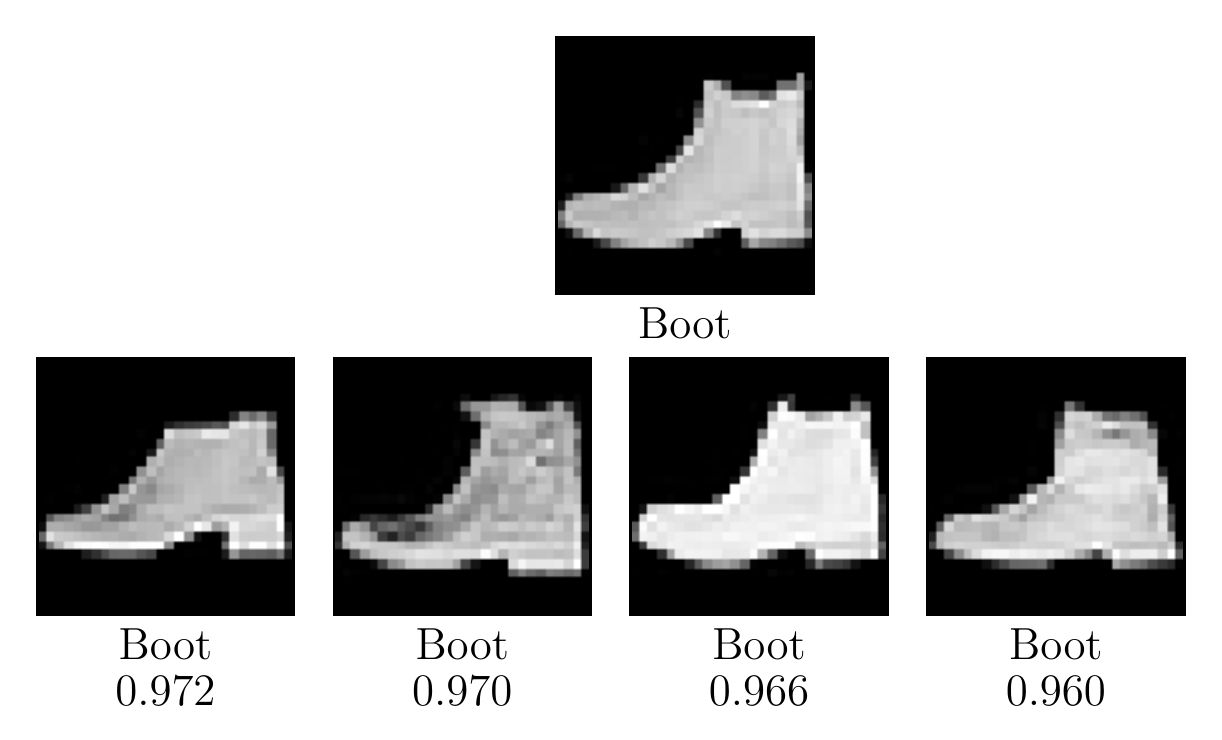}
\includegraphics[scale=0.6]{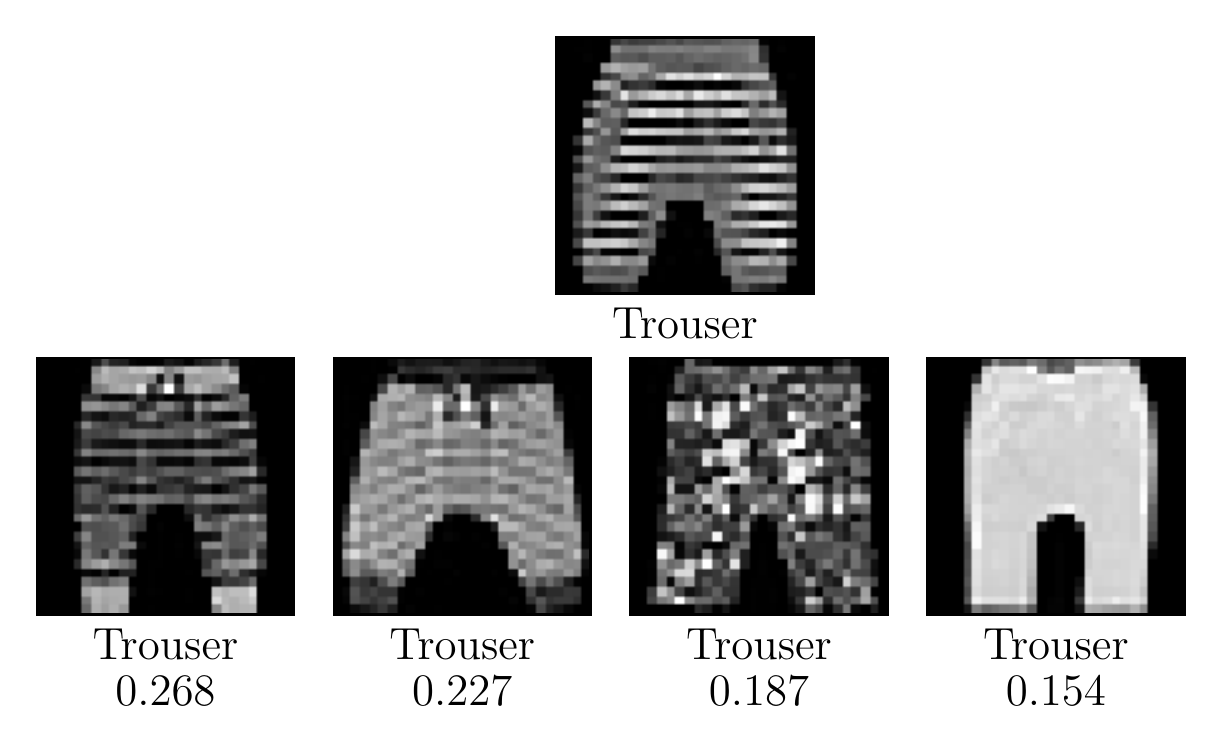}
\caption{Two examples of predictions made by the DWAC model on the Fashion MNIST dataset. An approximate explanation for the model's prediction on the test image (single image) is the four most-highly weighted training images (row of four), along with their weights.}
\label{fig:fashion}
\end{figure}

Recall that explanations for predictions made by our model are given in terms of a weighted sum of training instances.
Figure \ref{fig:fashion} (top) shows an example of a partial explanation provided by our model for an image dataset. A test image is shown in the top row, along with its true label, and the four closest images from the proper training set (as measured in the embedded space) are shown below it, along with their weights. In this case, all share the same label, and all contribute approximately equally to the prediction.

As a contrasting example, Figure \ref{fig:fashion} (bottom) shows an example which has poor support among the nearest training points. Although the  prediction is correct, and the closest training images appear visually similar, this sort of wide-legged trouser is quite rare in the dataset (most trousers included have narrow legs). As such, the model has clearly not learned as well how to embed such images into the low-dimensional space. The low weights indicate that we should be skeptical of this prediction. 

Images are relatively easy to compare at a glance; text, by contrast, may be more difficult. Nevertheless, the explanations given by our model for the predictions on text data are in some cases very meaningful. For the Stack Overflow dataset, for example, many instances are almost trivially easy, in that the label is part of the question. Not surprisingly, these examples tend to have many highly-weighted neighbors which provide a convincing explanation. Such an example is shown in Table \ref{tab:stackoverflow} (top). In other cases, the text is more ambiguous. Table \ref{tab:stackoverflow} (bottom) shows an example with very little support, for which a user might rightly be skeptical of the model, based on both the weights and the explanation.

\begin{table}
\centering
\begin{tabular}{ m{1.2cm} m{4.9cm} m{0.8cm}}
Weight & Sentence & Label \\
\hline
Test  & Drupal 6 dynamic menu item & Drupal \\
\hline
0.994 & drupal how to limit the display results of the calendar view & Drupal \\
0.994 & Drupal : Different output for first item in a block & Drupal \\
0.994 & changing user role in drupal  & Drupal \\
\end{tabular}
\newline
\vspace*{0.4 cm}
\newline
\begin{tabular}{ m{1.2cm} m{4.9cm} m{0.8cm}}
Weight & Sentence & Label \\
\hline
Test  & save data from editable division & Ajax \\
\hline
0.214 & Pass data from workspace to a function & Matlab \\
0.133 & upload data from excel to access using java & Apache \\
0.130 & Finding incorrectly - formatted email addresses in a CSV file & Excel \\ \\
\end{tabular}
\caption{Two examples from the Stack Overflow dataset with approximate explanations from a DWAC model: an easy example with many close neighbours (top) and a more difficult example with no close neighbours (bottom). The first line is the test instance in both cases.}
\label{tab:stackoverflow}
\end{table}

Finally, because we can use any deep model to compute $f(\mathbf{x})$, we are free to choose one with characteristics we prefer, including interpretability. For our text classification experiments, we chose a base model originally proposed for interpretable classification of medical texts \cite{mullenbach.2018}. To further unpack the explanation given for a prediction, we could, for example, inspect the attention weights for a particular pair of sentences to understand the importance of each word in context. 

For a comparable pair of examples from a tabular dataset, please refer to supplementary material.

\subsection{Approximate Explanations} \label{sec:approx}

Because the weights on training instances decrease exponentially with distance, the closest training instances will contribute the most to the prediction. In some cases, only relatively few training instances will be required to fully determine the model's predicted label (because beyond this the remaining instances will lack sufficient weight to alter which class will be most highly-weighted). In practice, most test instances tend to require a substantial proportion of the nearest training instances in order to cross this threshold. However, even considering a much smaller number of the closest training instances may still result in high agreement with the prediction based on all of the data.
Table \ref{tab:size} shows the agreement with the full model if we only consider the top $k$ neighbors to each test instance. For most datasets, this agreement is very high, even for a very small number of neighrbors.

\begin{table}
\centering
\begin{tabular}{l c c c c}
Dataset & k=1 & k=5 & k=10 & k=100 \\
\hline
Adult Income & 0.85 & 0.91 & 0.93 & 0.96 \\
Covertype & 0.76 & 0.77 &  0.77& 0.83 \\
Lending Club & 0.96 & 0.98 & 0.99 & 0.99 \\
Fashion MNIST & 0.95 & 0.97 & 0.98 & 0.99 \\
CIFAR-10  & 0.96 & 0.98 & 0.98 & 0.99 \\
Subjectivity & 0.98 & 0.99 & 0.99 & 1.00 \\
Stack Overflow & 0.81 & 0.84 & 0.85 & 0.89 \\
IMDB & 0.95 & 0.98 & 0.98 & 0.99 \\
Amazon & 0.78 & 0.87 & 0.90 & 0.94 \\
\end{tabular}
\caption{Impact of considering a subset of the training instances as an approximate explanation: the columns show agreement with the full model on the single most-probable label when basing the prediction on only the $k$ closest training instances. }
\label{tab:size}
\end{table}

\subsection{Confidence and Credibility} \label{sec:conformal_results}

\begin{figure}
\includegraphics[scale=0.55]{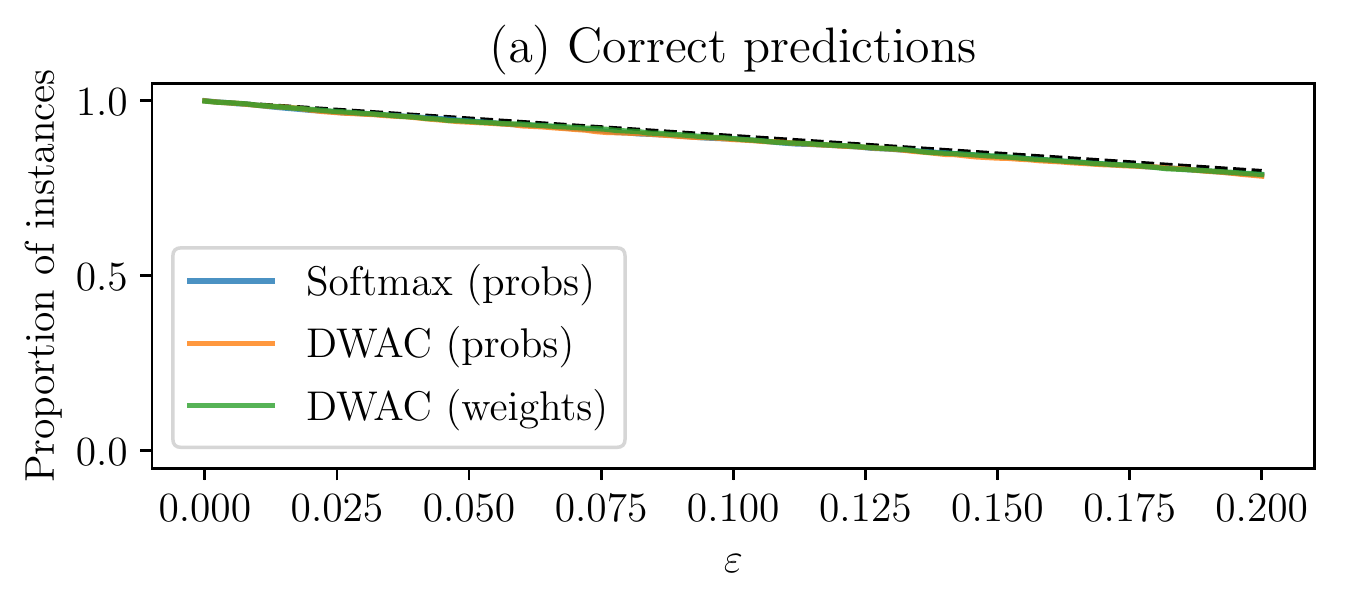}
\includegraphics[scale=0.55]{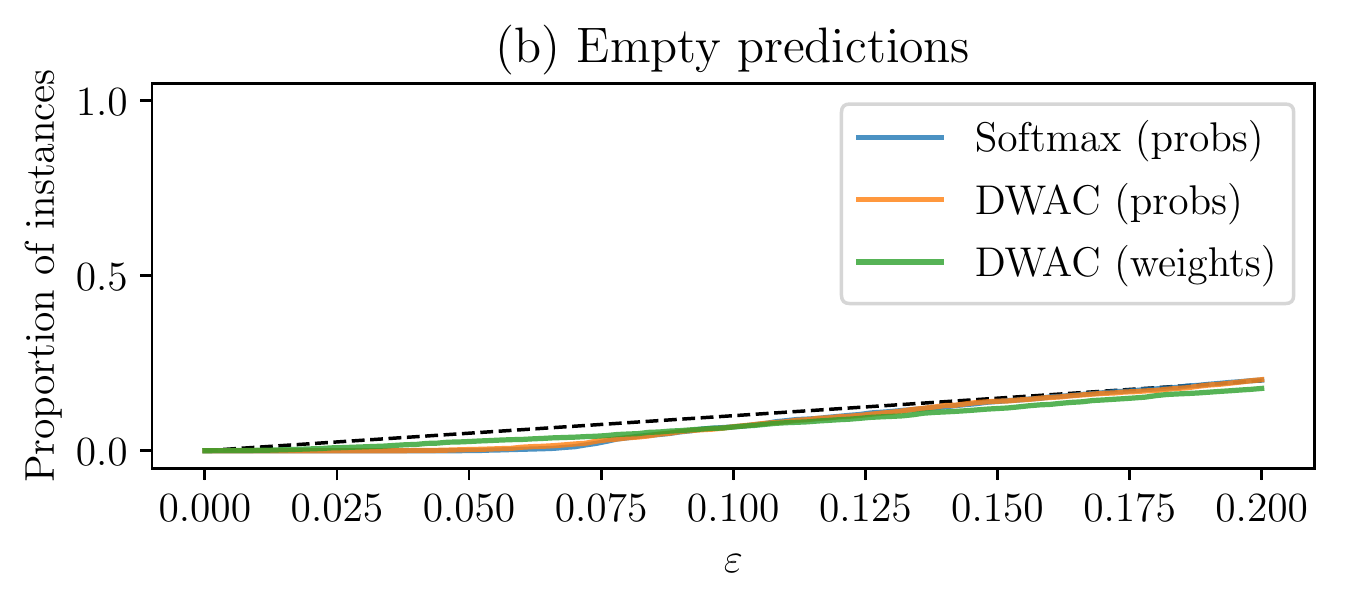}
\includegraphics[scale=0.55]{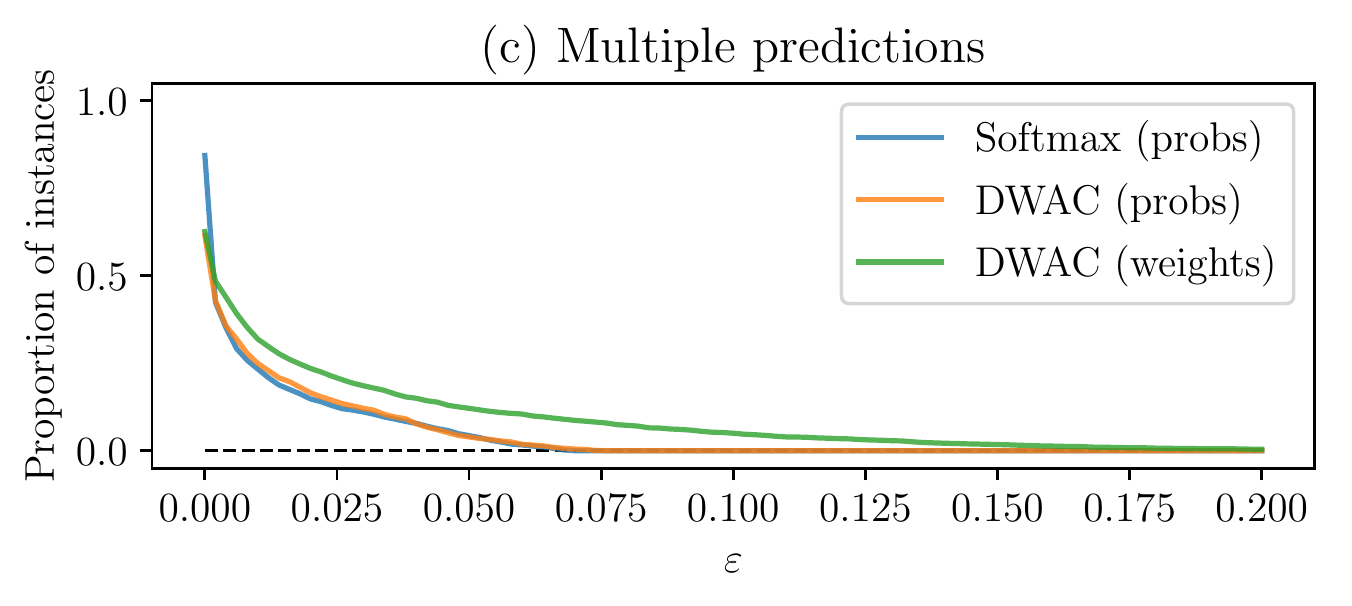}
\includegraphics[scale=0.55]{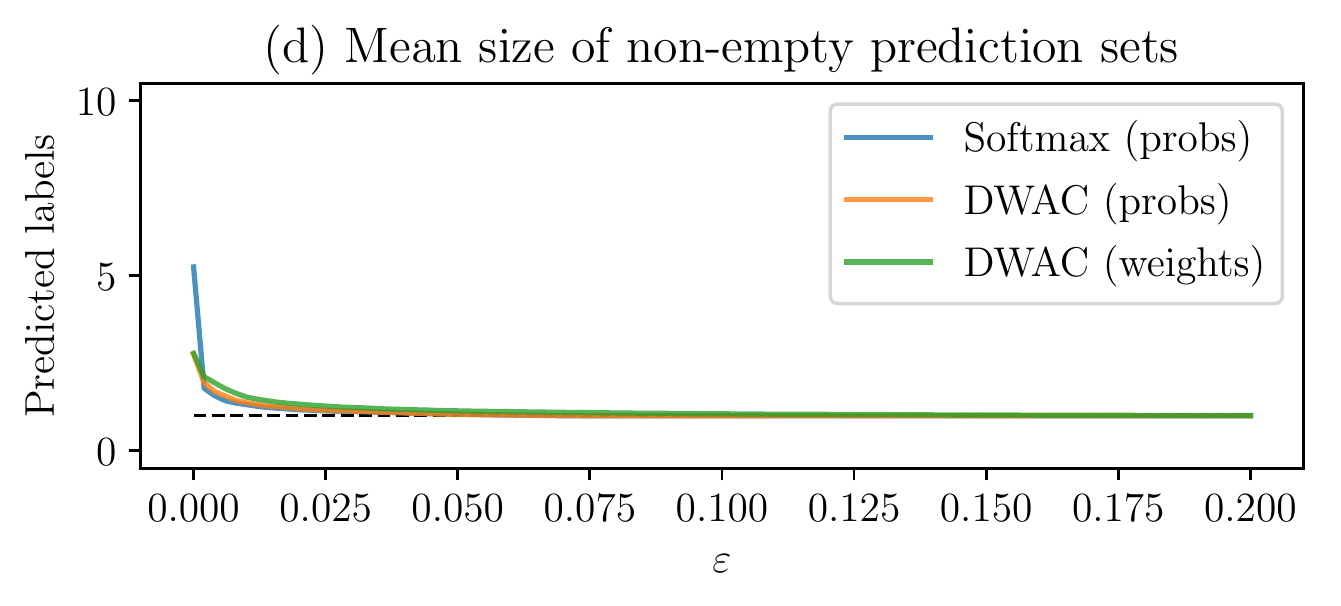}
\caption{Coverage of various models on the Fashion MNIST test data, as we vary the desired maximum error rate ($\varepsilon$). From top to bottom, the subfigures show (a) the proportion of predicted label sets that are correct (contain the true label); (b) that are empty (make no prediction); (c) that contain multiple labels; and (d) the mean number of labels in non-empty prediction sets. The softmax and DWAC models give nearly identical results when using negative probability as a measure of nonconformity (probs). Our proposed measure (weights) has an indistinguishable error rate, but is slightly less efficient. The dashed line in each figure represents an optimal response.}
\label{fig:fashion_conf}
\end{figure}

The above results only considered the single top-scoring label predicted for each instance; here we extend both models to the conformal setting. To verify the theoretical expectations of conformal methods, we show that our proposed measure of nonconformity correctly works to maintain a desired error rate in making predictions. Figure \ref{fig:fashion_conf} shows the results for the Fashion MNIST dataset, where we vary $\varepsilon$ (the desired maximum error rate) from 0 to 0.2.
The top subfigure (a) shows the proportion of predictions on the test set which are \textit{correct} (that is, which contain the true label), for the softmax model using negative probability (probs) as a measure of nonconformity, our model using the same measure, and our model using our proposed measure of nonconformity (weights) given in equation (\ref{eq:measure}).
As can be seen, all three demonstrate correct coverage, with all lines close to but not exceeding the expected proportion ($1-\varepsilon$) across the full range (shown as a dashed line on the top subfigure).
Note that this is not the same as accuracy, as some predictions may contain multiple labels.

The second and third subfigures show the same lines for the proportion of predicted label sets that are empty (b) or contain multiple labels (c). The bottom figure shows the mean number of labels in all non-empty predicted label sets. In all cases, the dashed line represents an optimal outcome (that is, a proportion of predicted label sets equal to $\varepsilon$ are empty, all other predictions are correct, and no predictions contain multiple labels).

As can be seen, the softmax and DWAC models give indistinguishable results when using the same measure of nonconformity. Our proposed measure of nonconformity, by contrast, appears to be slightly less efficient, producing slightly more predicted label sets with multiple labels, but also slightly more empty label sets, which represent identifiable errors contributing to the proportion of incorrect predictions.

The advantage of our proposed measure, however, comes in robustness to out-of-domain data. If we train a model on the Fashion MNIST dataset, and then ask it to make predictions on the original MNIST digits dataset (which has the same size and data format, but consists of hand-written digits rather than items of clothing), we would hope that a good model would predict relatively low probability for all such instances.
In fact, as has been previously observed \cite{nguyen.2015.deep}, deep models tend to predict relatively high probabilities,  even for out-of-domain data, and this is also true of DWAC models.

\begin{figure}
\includegraphics[scale=0.55]{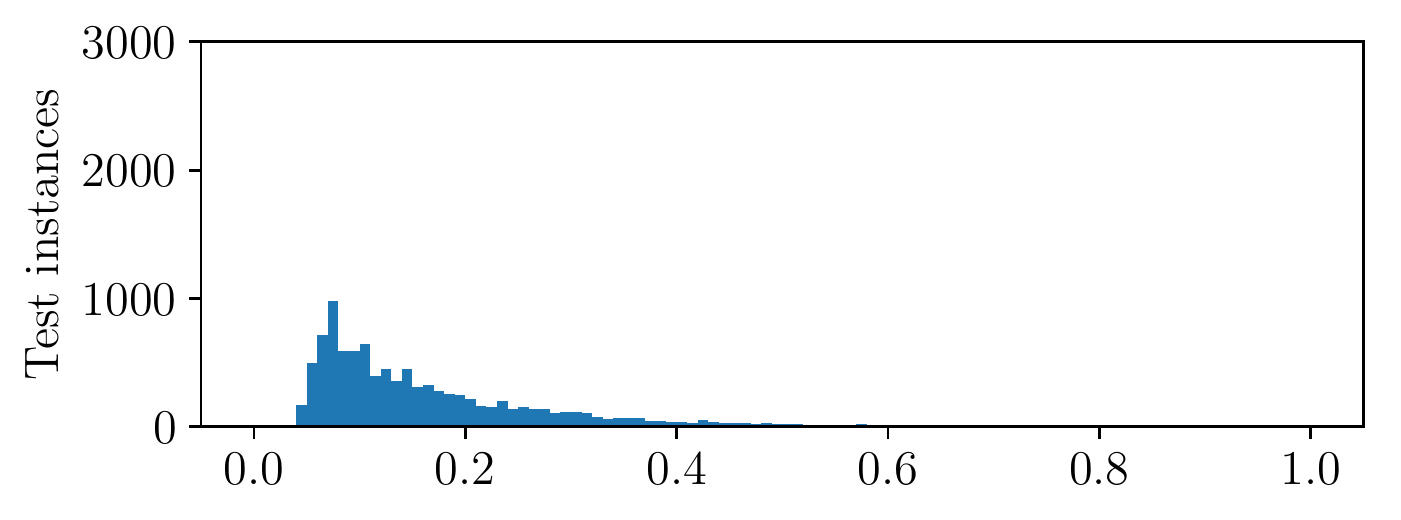}
\includegraphics[scale=0.55]{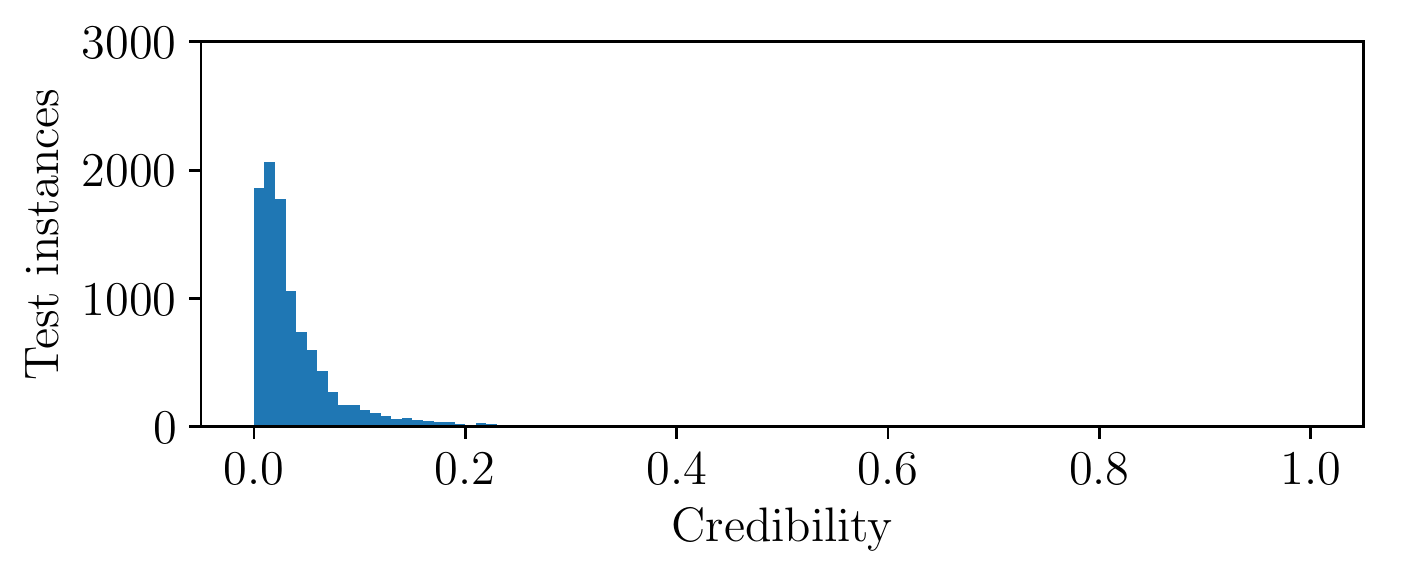}
\caption{Empirical distribution of credibility scores from the softmax (top) and DWAC (bottom) models when trained on Fashion MNIST and tested on MNIST digits (which have the same input format but different content), with the latter using our proposed measure of nonconformity.}
\label{fig:ood_creds}
\end{figure}

Fortunately, the \textit{credibility} score from a conformal predictor provides a meaningful estimate of how much we should trust the corresponding prediction.\footnote{Specifically, as mentioned above, it is equal to one minus the model probability that none of the labels should be given to this instance.} Both the softmax model (using negative probability as a measure of nonconformity), and the DWAC model (using our proposed measure of nonconformity) give low credibility to the vast majority of out-of-domain examples, as shown in Figure \ref{fig:ood_creds}. The credibility scores from DWAC however, are noticeably shifted closer to zero, indicating that the sum of the weights of the corresponding class is a better measure when we are concerned about the possibility of out-of-domain data. (For in-domain data, the credibility values will be approximately uniformly distributed).

We see a similar result when training on CIFAR-10 images, and predicting on the Tiny Images dataset, and an even more extreme difference in the case of the Covertype dataset, where we treat one out of seven classes as out-of-domain data, and train a model on only the six remaining classes. For that setup, the mean credibility score from the softmax model on the out-of-domain data is 0.9, whereas for the DWAC model with our measure of nonconformity it is 0.2 (figures in supplementary material). This indicates that to the softmax model, the held-out images ``look like'' even more extreme versions of one of the other classes, whereas the DWAC model correctly recognizes that the held-out images are relatively unlike the training instances (relative to the calibration data).


\section{Discussion and future work}

The idea of interpretability has always been important in statistics and machine learning, but has taken on a renewed urgency with the increased expressive power of deep learning models and the expanded deployment of machine learning systems in society. No single approach to interpretability is likely to solve all problems; here we have focused on adapting existing models so as to make their predictions more transparent, by decomposing them into a sum over training instances, the support for each of which can be inspected. As emphasized by papers that have made use of user studies \cite{huysmans.2011,kulesza.2013,narayanan.2018}, careful empirical work is required to evaluate the effectiveness of explanations, and we leave such an evaluation of this approach for future work.

A few recent papers have also sought to formalize the question of when to trust a classifier. Jiang et al.~\cite{jiang.2018} present a method for determining how much any existing classifier should be trusted on any given test point, based on the relative distance to the most-likely and second-most-likely clusters, with clusters based on a pruned training set. This appears to be a theoretically well-motivated and empirically effective technique, but is more focused on trust than interpretability. In an unpublished paper, Papernot and McDaniel~\cite{papernot.2018} propose to train a conventional deep model, but make predictions using a $k$-nearest neighbours approach, with distance computed using all internal nodes of the network. We, by contrast, propose to train a model using the same form as will be used for prediction, to make predictions based on a weighted sum over the entire training set, and rely on similarities computed in the low-dimensional space of the final layer. Wallace et al.~\cite{wallace.2018} apply the method from Papernot and McDaniel to the problem of text classification and explore the implications for interpretability.

There have also been several papers focused on the problem of predicting whether data is in-domain or out-of-domain \cite{lee.2018,liang.2018}.
Many of these build on Hendrycks and Gimpel \cite{hendrycks.2017}, who observed that the predicted probabilities contain some useful signal as to whether data came from in-domain or out-of-domain, and proposed to use this to differentiate between the two by thresholding these probabilities. The authors did not, however, make the connection to conformal methods, which offer a more theoretically sound basis on which to make these decisions, as well as greater flexibility of metrics, beyond just predicted probability.

There are several natural extensions to this work which could be pursued, such as applying a similar architecture to regression or multi-label problems, as well as extending the idea of nonconformity to provide class-conditional guarantees.



\section{Conclusions}

In this paper we have demonstrated that even for sophisticated deep learning models, it is possible to create a nearly identical model, with all of the same desirable properties, that nevertheless provides an explanation for any prediction in terms of a weighted sum of training instances. In domains where the training data can be freely inspected, this provides greater transparency by revealing the many components that are explicitly contributing to a model's prediction, each of which can in principle be inspected and interrogated. Moreover, this method can build on top of other approaches to interpretability, by choosing an appropriate base model. When an approximate explanation will suffice, then using only a small subset of the training instances provides a natural high-fidelity approximation. More importantly, representing the prediction in this manner suggests a natural alternative measure of nonconformity, which, as we have shown, provides a more effective measure for detecting out-of-domain examples. Even in cases where training data cannot be shared (for privacy reasons, for example), this use of conformal methods still allows us to assert a quantitative estimate of the credibility of an individual prediction, one that is far more meaningful than the model's predicted probability.

\newpage

\section*{Acknowledgments}
We acknowledge helpful comments from the anonymous reviewers.  This research was funded in part by a University of Washington
Innovation award and an REU supplement to NSF grant IIS-1562364.

\bibliographystyle{ACM-Reference-Format}
\bibliography{refs}

\pagebreak


\section*{Supplementary Material}

\renewcommand{\thesubsection}{\Alph{subsection}}

\subsection{Dataset Details}

\begin{itemize}
\item Adult Income: Used the standard train/test split; dropped the ``fnlwgt'' column; normalized continuous variables and and converted categorical variables to indicators.\footnote{\url{https://archive.ics.uci.edu/ml/datasets/adult}}
\item Covertype: Used a random train/test split; normalized continuous variables.\footnote{\url{https://archive.ics.uci.edu/ml/datasets/covertype}}
\item Lending Club: Used a random train/test split. Used an established preprocessing pipeline\footnote{\url{https://www.kaggle.com/wsogata/good-or-bad-loan-draft/notebook}}; defined the positive category as ``Current'', and the negative category as everything else but ``Fully Paid'' or ``Does not meet the credit policy''.\footnote{\url{https://www.kaggle.com/wendykan/lending-club-loan-data}}
\item Fashion MNIST: Used standard train/test split; applied image-level normalization.\footnote{\url{https://github.com/pytorch/vision/blob/master/torchvision/datasets/mnist.py}}
\item MNIST: Used the standard test set as out-of-domain for classifiers trained on Fashion MNIST; applied image-level normalization.\footnote{\url{https://github.com/pytorch/vision/blob/master/torchvision/datasets/mnist.py}}
\item CIFAR-10: Used the standard training/test split and an existing preprocessing pipeline.\footnote{\url{https://www.cs.toronto.edu/~kriz/cifar.html}}$^,$\footnote{\url{https://github.com/kuangliu/pytorch-cifar}}
\item Subjectivity: Used a random train/test split of the subjectivity dataset v1.0.\footnote{http://www.cs.cornell.edu/people/pabo/movie-review-data/}
\item TinyImages: Used the training images as an out-of-domain dataset for CIFAR-10.\footnote{http://horatio.cs.nyu.edu/mit/tiny/data/index.html}
\item Stack Overflow: Used a random train/test split.\footnote{\url{https://github.com/jacoxu/StackOverflow}}
\item IMDB: Used the standard train/test split.\footnote{\url{http://ai.stanford.edu/~amaas/data/sentiment/}}
\item Amazon: Used a random train/test split of the 5-core subset of the ``Beauty'' category.\footnote{\url{http://jmcauley.ucsd.edu/data/amazon/}}
\item For all text datasets, text was tokenized with spaCy,\footnote{\url{https://spacy.io/}} and converted to lower case, using a vocabulary built from the training set, with word embeddings initialized using 300-dimensional Glove vectors trained on the 6 billion tokens of Wikipedia 2014 and Gigaword 5.\footnote{\url{https://nlp.stanford.edu/projects/glove/}}
\end{itemize}

\subsection{Model details}

\begin{itemize}
\item Tabular model: All tabular datasets used a multi-layer perceptron with layers of size $(|\mathbf{x}|, 32, 8, |\mathbf{h}|)$, with dropout ($p=0.2$) after each hidden layer.
\item Fashion MNIST model: We use a conventional architecture with two convolution layers with kernel size 3, first with 32 filters, then 64 filters, followed by max pooling and dropout $(p=0.25)$, followed by a nonlinear layer mapping from 9216 dimensions to 128, followed by dropout ($p=0.5$), followed by a linear projection.
\item CIFAR-10 model: We use an implementation of ResNet18, but get somewhat worse that reported performance for the basic model, possibly due to using a fixed learning rate.\footnote{\url{https://github.com/kuangliu/pytorch-cifar}}
\item Text model: Convolutional model with attention, with an initial embedding layer (initialized using glove vectors), followed by dropout ($p=0.5$), followed by a convolution with window size 5 and 200 filters, followed by an attention-weighted sum, followed by a linear projection layer.
\end{itemize}

\subsection{Additional Results}

\subsubsection{Amazon Product Reviews}

\begin{figure}
\includegraphics[scale=0.55]{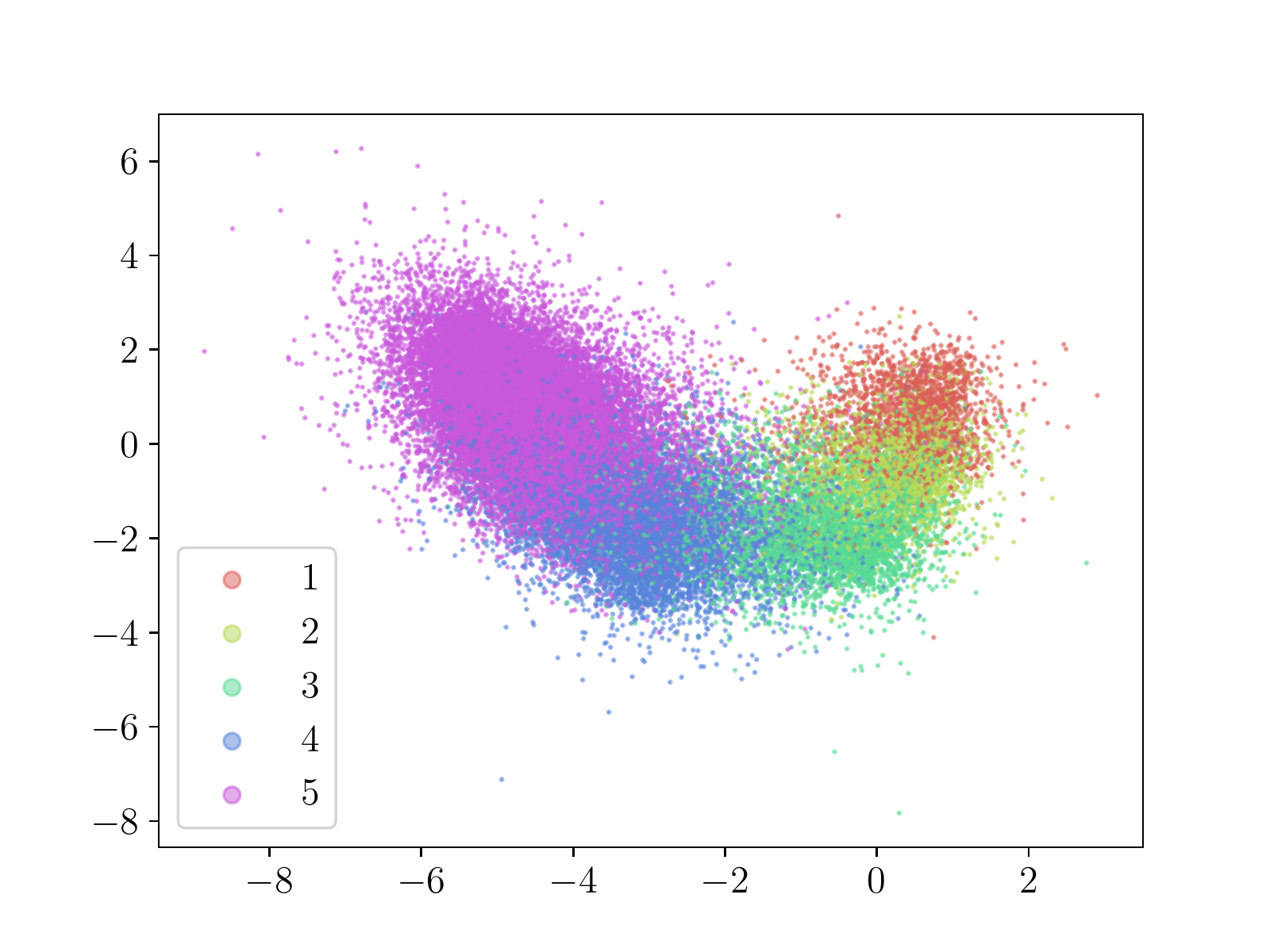}
\caption{Learned embeddings of the Amazon product reviews training data when using a DWAC model with a two-dimensional output layer.}
\label{fig:amazon_scatter}
\end{figure}

Similar to Figure 2 in the main paper, if we train a DWAC model with a two-dimensional output layer on the Amazon product reviews, each labeled as 1-5 stars, the embeddings have a natural semantics that gradually transitions from 1 to 5, even though the model treats these as unordered categories. An example is shown in Figure \ref{fig:amazon_scatter}.

\subsubsection{Adult Income Data}

Table \ref{tab:income} shows two examples from the Adult Income dataset of the kind of approximate explanation that might be offered by the DWAC model. One one level, the two test instances are similar, in that they both young, poorly-educated, and working in private industry. However, considering the embedded representation, the first (top) has many similar training instances, resulting in a high-credibility prediction, whereas the second (bottom) is an unusual instance for which there appear to be few similar cases in the training data, resulting in a low-credibility prediction.


\begin{table*}
\centering
\begin{tabular}{l c c c c c c}
Instance & Test & Train1 & Train2 & Train3 & Train4 & Train5 \\
Label & $\leq 50k$ & $\leq 50k$ & $\leq 50k$ & $\leq 50k$ &$\leq 50k$ & $\leq 50k$ \\
Weight & n/a & 0.999 & 0.999 & 0.999 & 0.999 & 0.999 \\
\hline 
Age & 31 & 31 & 30 & 29 & 30 & 44 \\
Work &  Private & Private & Private & Local-gov & Private & Private \\
Education & HS-grad & HS-grad & Some college & HS-grad & Some college & Some college \\
Education years & 9 & 9 & 10 & 9 & 10 & 10 \\
Martial status & Never married  & Never married & Never married &Never married & Divorced & Separated \\
Occupation & Transport & Transport & Craft-repair & Protective-serv & Transport & Adm-clerical \\
Relationship & Not-in-family & Not-in-family & Not-in-family & Not-in-family & Not-in-family & Unmarried \\
Race & White & White & Black & White & White & White \\
Sex & Male & Male & Male & Male & Male &  Female \\
Capital Gains & 0  & 0  & 0  & 0  & 0  & 0 \\
Capital Losses & 0  & 0  & 0  & 0  & 0  & 0 \\
Hours per week & 50 & 50 & 45 & 40 & 40 & 40 \\
Country & USA & USA & USA  & USA  & USA  & USA
\end{tabular}
\newline
\vspace*{0.4 cm}
\newline
\centering
\begin{tabular}{l c c c c c c}
Instance & Test & Train1 & Train2 & Train3 & Train4 & Train5 \\
Label & $> 50k$ & $> 50k$ & $> 50k$ & $> 50k$ &$> 50k$ & $> 50k$ \\
Weight & n/a & 0.883 & 0.816 & 0.695 & 0.608 & 0.460 \\
\hline 
Age & 30 & 55 & 30 & 56 & 22 & 33 \\
Work &  Private & Private & Self-emp-not-inc & Self-emp-not-inc & Self-emp-not-inc & Private \\
Education & HS-grad & $9^\textrm{th}$  & HS-grad & HS-grad & Some college & Some college \\
Education years & 9 & 5 & 9 & 9 & 10 & 10 \\
Martial status & Never married  & Divorced & Never married & Widowed & Never married & Never married \\
Occupation & Handlers-cleaners & Craft-repair & Craft-repair & Adm-clerical & Sales & Adm-clerical \\
Relationship & Own-child & Unmarried & Not-in-family & Unmarried  & Own-child & Own-child \\
Race & Black & White & White & White & Black & White \\
Sex & Female & Female & Male & Female & Male &  Female \\
Capital Gains & 99999 & 99999 & 99999 & 99999 & 99999 & 99999 \\
Capital Losses & 0  & 0  & 0  & 0  & 0  & 0 \\
Hours per week & 40 & 37 & 50 & 40 & 55 & 30 \\
Country & USA & USA & USA  & USA  & USA  & USA
\end{tabular}
\caption{Examples of a person that is being predicted to make less than \$50,000 per year, with high probability, and high credibility, along with the most similar training instances (top); and a person being predicted to make more than \$50,000 per year, with high probability but low credibility, using the DWAC model (bottom).}
\label{tab:income}
\end{table*}

\subsubsection{Credibility values}

\begin{figure}
\includegraphics[scale=0.55]{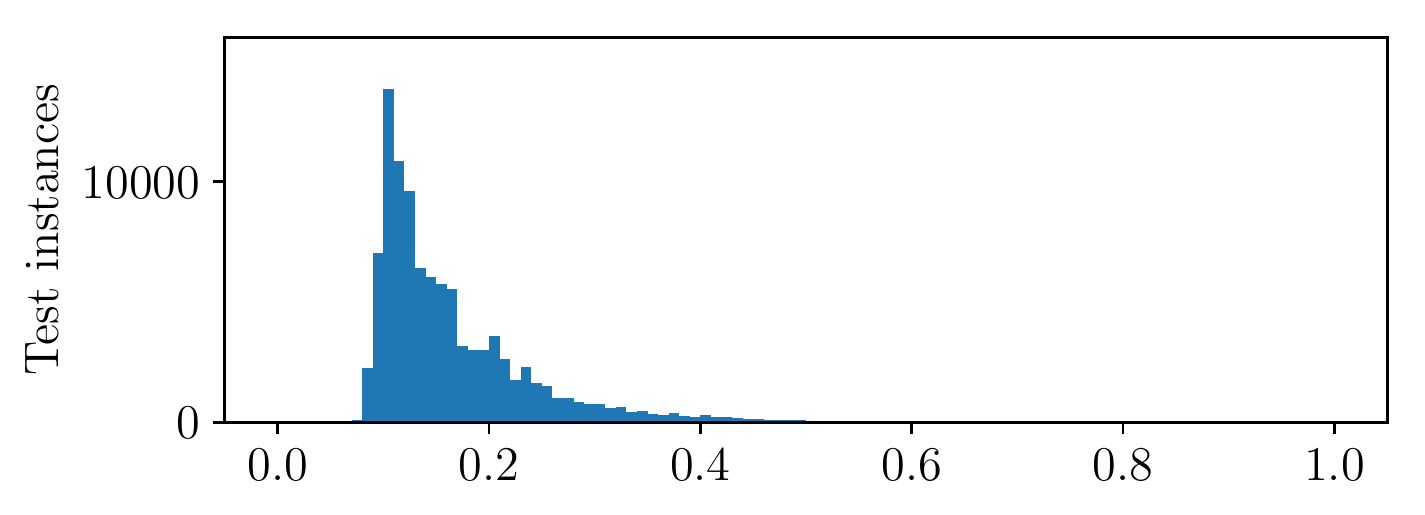}
\includegraphics[scale=0.55]{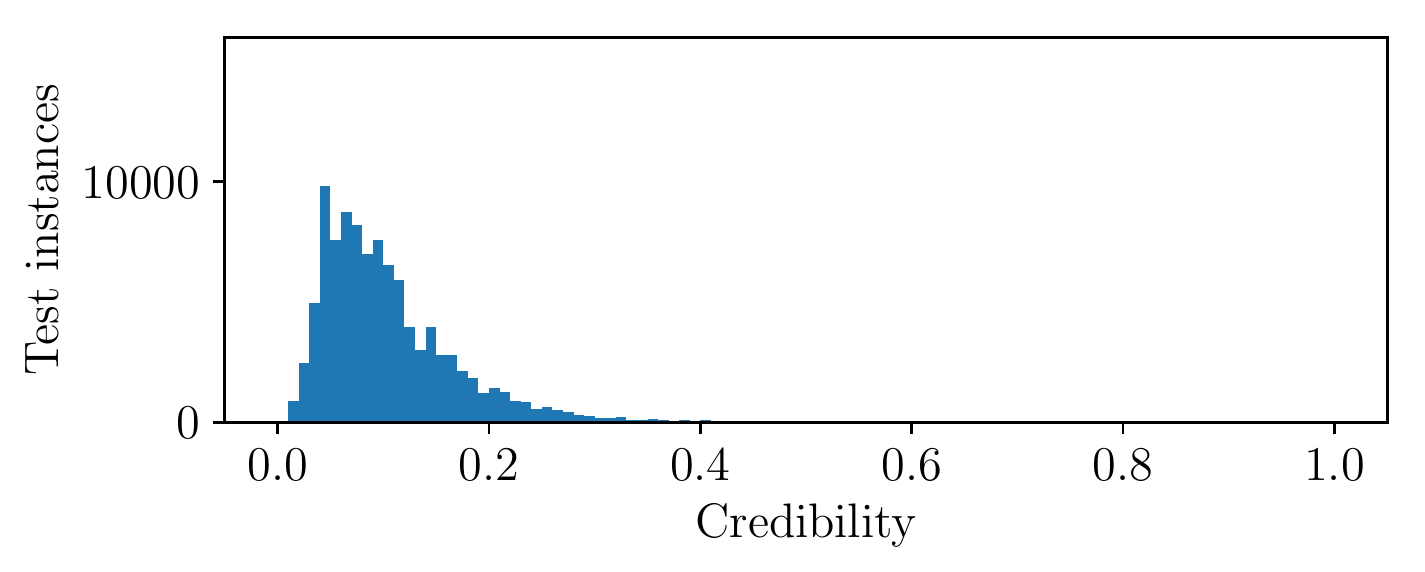}
\caption{Empirical distribution of credibility scores from the softmax (top) and DWAC (bottom) models when trained on CIFAR-10 and tested on Tiny Images, with the latter using our proposed measure of nonconformity.}
\label{fig:cifar_ood}
\end{figure}

\begin{figure}
\includegraphics[scale=0.55]{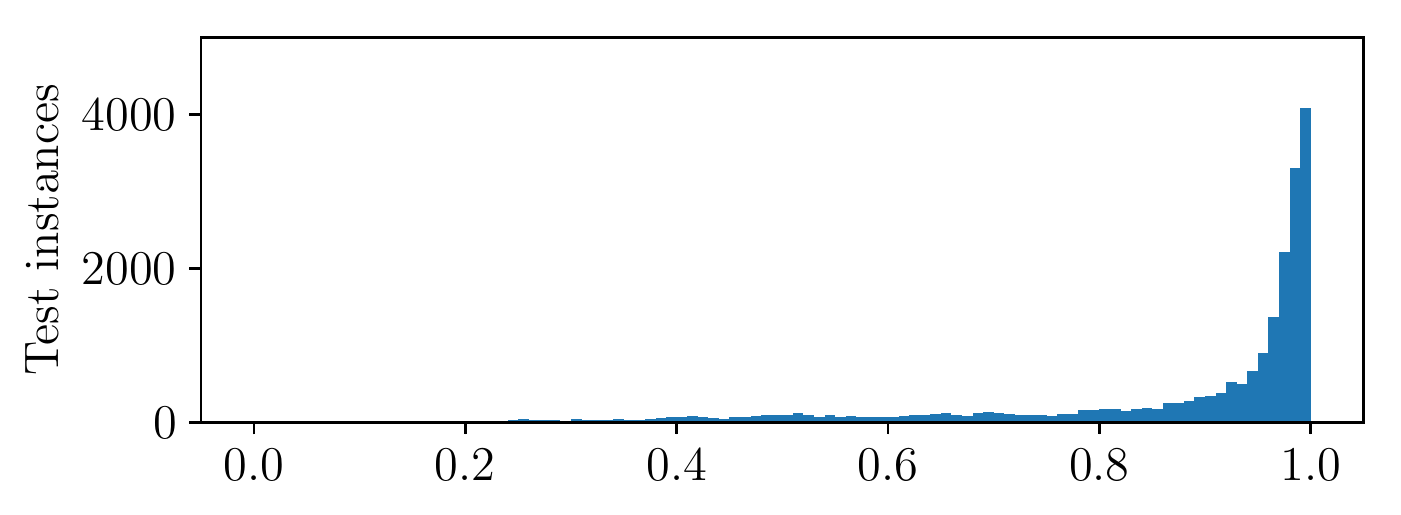}
\includegraphics[scale=0.55]{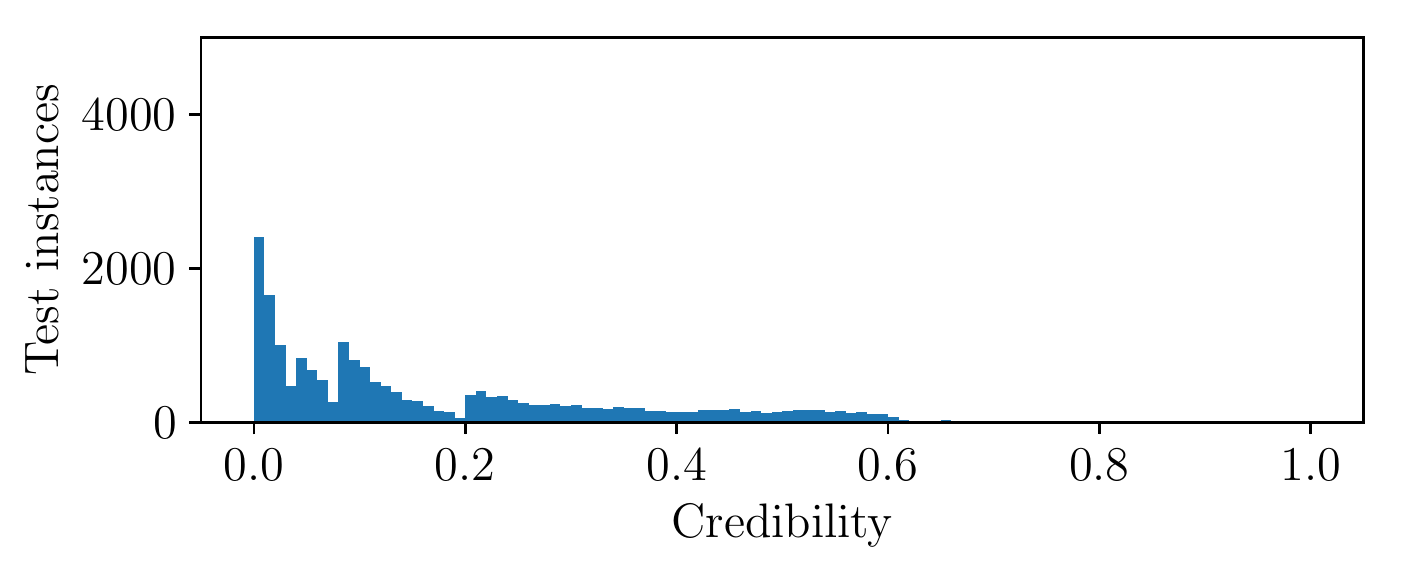}
\caption{Empirical distribution of credibility scores from the softmax (top) and DWAC (bottom) models when trained on 6 of the classes in the Covertype dataset and tested on the 7th class. This represents an extreme example where the credibility scores are skewed towards 1 in the softmax model, given that in-domain data would typically be approximately uniformly distributed.}
\label{fig:covertype_ood}
\end{figure}

Figure \ref{fig:cifar_ood} shows the plots of credibility values from the softmax and DWAC models when trained on CIFAR-10 and asked to predict on the Tiny Images dataset. Figure \ref{fig:covertype_ood} shows the same thing for the Covertype dataset when one class (``Krummholz'') is held out as out-of-domain, and the models are only trained on data representing the six remaining classes. In the first case, the models give quite similar distributions of credibility values, whereas in the latter the DWAC model is much better at detecting that the examples given are out-of-domain. Both models predict that the most likely label for almost all of the out-of-domain instances is ``Spruce/Fir'', the second most common class. However, the DWAC model recognizes that these do not appear to be typical training instances, whereas the softmax model reports that they appear to be even more like a particular class than is typical among calibration instances.

\end{document}